%% file: main.tex
\documentclass{article}
\usepackage{spconf}
\input{math_commands.tex}

\usepackage{hyperref}
\usepackage{easy-todo}
\usepackage{graphicx}
\usepackage{amsmath,amssymb,amsthm}
\usepackage{tikz}
\usepackage{forloop}
\usepackage{makecell}
\usepackage[ruled, linesnumbered]{algorithm2e}

\def\dncl{d_{\mathrm{ncl}}}

\title{Dynamic Texture Recognition via Nuclear Distances on Kernelized Scattering Histogram Spaces}
\name{Alexander Sagel, \quad Julian W{\"o}rmann, \quad Hao Shen\sthanks{\texttt{\{sagel,woermann,shen\}@fortiss.org}}}
\address{fortiss - The Research Institute of the Free State of Bavaria\\Munich, Germany}
\begin{document}

%
\maketitle
\begin{abstract}
Distance-based dynamic texture recognition is an important research field in multimedia processing with applications ranging from retrieval to segmentation of video data. Based on the conjecture that the most distinctive characteristic of a dynamic texture is the appearance of its individual frames, this work proposes to describe dynamic textures as kernelized spaces of frame-wise feature vectors computed using the Scattering transform. By combining these spaces with a basis-invariant metric, we get a framework that produces competitive results for nearest neighbor classification and state-of-the-art results for \emph{nearest class center} classification.
\end{abstract}
\section{Introduction}
The term  \emph{temporal} or \emph{dynamic texture} refers to a class of visual processes consisting of a temporally evolving image texture. Typical examples of dynamic textures are ocean waves, flags moving in the wind, or flames of a bonfire.
In order to classify, retrieve or cluster dynamic textures, but also for purposes such as spatial or temporal video segmentation or identifying patterns in a dynamic scene, an expressive way to represent dynamic textures and to measure similarity between them is of great help. 
Since a dynamic texture is defined by the appearance of its individual frames, as well as the dynamics of their temporal progression, 
descriptors usually aim at capturing the behavior 
along the spatial \emph{and} temporal axes.

Quite often, the dynamics are of considerably lesser importance for distinguishing videos of temporal textures than the appearance of individual frames. For instance, to tell apart a forest from an ocean, we do not need to consider, whether the video was shot in stormy or calm weather conditions. In fact, dynamic textures can be usually easily recognized by observing isolated video frames. In this work, we therefore neglect the issue of appropriately describing the temporal transitions. Instead, we treat each dynamic texture as a set of image feature vectors describing individual frames. We then proceed to model each set as a finite subspace of a kernel \emph{feature space} that is represented as a coefficient matrix of an orthogonal basis. Since one vector space is spanned by an infinite number of orthogonal bases, we employ a basis-invariant metric.

This work incorporates two observations from previous publications. The first observation is that when we apply 
\emph{Scattering transform} to a texture image, 
its
coefficient distributions
constitute an expressive feature representation that performs well on recognition tasks when combined with probability product kernels \cite{Sagel2015}. The second observation is that visual processes can be well described as spaces computed by applying Kernel PCA \cite{Schoelkopf1998} on distribution-based descriptors of the video frames, when employed together with a distance measure that is invariant to the bases of said spaces \cite{Sagel2017, Chaudhry2009}. The contributions of this work are as follows. First, we propose a feature extraction method that performs a Scattering transform on an image texture and computes a histogram from each Scattering \emph{subband}. These histograms are then concatenated into a \emph{Scattering histogram vector} and a Mercer kernel on pairs of these vectors is defined. Second, we introduce an algorithm which expects a dynamic texture video and returns a coefficient matrix that describes the kernel subspace containing the Scattering histogram vectors computed from the video. Next, we define a metric on pairs of these coefficient matrices. 
Finally, we describe an algorithm for computing \emph{Fr{\'e}chet means} \cite{Sagel2017} on finite sets of coefficient matrices with respect to the defined metric. 

The classical approach to \emph{dynamic texture recognition} is by first computing a linear autoregressive state-space model, and then defining a distance measure \cite{Saisan2001} on its parameters. Another common approach is to use multi-scale spatio-temporal filter responses of the dynamic texture video, e.g. the \emph{2D-T Curvelet Transform} \cite{Dubois2015} or the \emph{Spatio-temporal receptive fields} \cite{Jansson2018}. Beyond that, a considerable number of recognition methods in the recent years is based on the idea to collect features that are designed for 2D images from a video, by applying them in \emph{three orthogonal planes} (TOP) of the video cuboid  \cite{Zhao2007, Nguyen2017}. Methods based on computing sparse representations have also demonstrated remarkable performance \cite{Quan2015}. 
Interestingly, while neural networks play a significant role in visual process recognition nowadays \cite{Qi2016}, classical end-to-end
deep learning is more an exception than the rule.
One reason for this might be that even today it is not straight-forward to collect enough dynamic textures sequences to train a deep neural net with a superior performance.
\section{Scattering Subband Histograms}
\begin{figure}
	\begin{center}
		\begin{tikzpicture}[scale=0.17]
		\node[font=\small] at (24,19.5) {$x$};
		\draw[->] (24,19) -- (24,18);
		\draw[fill=white, rounded corners] (22.5,16.6) rectangle (25.5,18);
		\node[font=\scriptsize] at (24,17.3) {$\Psi$};
		\draw[->] (22.5,17.3) -- (21.5,17.3);
		\draw[fill=white] (21,17.3) circle (0.5);
		\draw[fill=black] (8,14.5) circle (0.5);
		\draw[->] (23,16.6) -- (8.495,14.569);
		\draw[->] (8,14) -- (8,13);
		\draw[fill=white, rounded corners] (6.5,11.6) rectangle (9.5,13);
		\node[font=\scriptsize] at (8,12.3) {$\Psi$};
		\draw[->] (6.5,12.3) -- (5.5,12.3);
		\draw[fill=white] (5,12.3) circle (0.5);
		\draw[fill=black] (2.667,9.5) circle (0.5);
		\draw[->] (7,11.6) -- (3.132,9.683);
		\draw[->] (2.667,9) -- (2.667,8);
		\draw[fill=white, rounded corners] (1.167,6.6) rectangle (4.167,8);
		\node[font=\scriptsize] at (2.667,7.3) {$\Psi$};
		\draw[->] (1.167,7.3) -- (0.167,7.3);
		\draw[fill=white] (-0.333,7.3) circle (0.5);
		\draw[fill=black] (0.889,4.5) circle (0.5);
		\draw[->] (1.667,6.6) -- (1.063,4.969);
		\draw[fill=black] (2.667,4.5) circle (0.5);
		\draw[->] (2.667,6.6) -- (2.667,5);
		\draw[fill=black] (4.444,4.5) circle (0.5);
		\draw[->] (3.667,6.6) -- (4.271,4.969);
		\draw[fill=black] (8,9.5) circle (0.5);
		\draw[->] (8,11.6) -- (8,10);
		\draw[->] (8,9) -- (8,8);
		\draw[fill=white, rounded corners] (6.5,6.6) rectangle (9.5,8);
		\node[font=\scriptsize] at (8,7.3) {$\Psi$};
		\draw[->] (6.5,7.3) -- (5.5,7.3);
		\draw[fill=white] (5,7.3) circle (0.5);
		\draw[fill=black] (6.222,4.5) circle (0.5);
		\draw[->] (7,6.6) -- (6.396,4.969);
		\draw[fill=black] (8,4.5) circle (0.5);
		\draw[->] (8,6.6) -- (8,5);
		\draw[fill=black] (9.778,4.5) circle (0.5);
		\draw[->] (9,6.6) -- (9.604,4.969);
		\draw[fill=black] (13.333,9.5) circle (0.5);
		\draw[->] (9,11.6) -- (12.869,9.683);
		\draw[->] (13.333,9) -- (13.333,8);
		\draw[fill=white, rounded corners] (11.833,6.6) rectangle (14.833,8);
		\node[font=\scriptsize] at (13.333,7.3) {$\Psi$};
		\draw[->] (11.833,7.3) -- (10.833,7.3);
		\draw[fill=white] (10.333,7.3) circle (0.5);    
		\draw[fill=black] (11.555,4.5) circle (0.5);
		\draw[->] (12.333,6.6) -- (11.729,4.969);
		\draw[fill=black] (13.333,4.5) circle (0.5);
		\draw[->] (13.333,6.6) -- (13.333,5);
		\draw[fill=black] (15.111,4.5) circle (0.5);
		\draw[->] (14.333,6.6) -- (14.937,4.969);
		\draw[fill=black] (24,14.5) circle (0.5);
		\draw[->] (24,16.6) -- (24,15);
		\draw[->] (24,14) -- (24,13);
		\draw[fill=white, rounded corners] (22.5,11.6) rectangle (25.5,13);
		\node[font=\scriptsize] at (24,12.3) {$\Psi$};
		\draw[->] (22.5,12.3) -- (21.5,12.3);
		\draw[fill=white] (21,12.3) circle (0.5);
		\draw[fill=black] (18.667,9.5) circle (0.5);
		\draw[->] (23,11.6) -- (19.132,9.683);
		\draw[->] (18.667,9) -- (18.667,8);
		\draw[fill=white, rounded corners] (17.167,6.6) rectangle (20.167,8);
		\node[font=\scriptsize] at (18.667,7.3) {$\Psi$};
		\draw[->] (17.167,7.3) -- (16.167,7.3);
		\draw[fill=white] (15.667,7.3) circle (0.5);    
		\draw[fill=black] (16.889,4.5) circle (0.5);
		\draw[->] (17.667,6.6) -- (17.063,5);
		\draw[fill=black] (18.667,4.5) circle (0.5);
		\draw[->] (18.667,6.6) -- (18.667,5);
		\draw[fill=black] (20.444,4.5) circle (0.5);
		\draw[->] (19.667,6.6) -- (20.27,5);
		\draw[fill=black] (24,9.5) circle (0.5);
		\draw[->] (24,11.6) -- (24,10);
		\draw[->] (24,9) -- (24,8);
		\draw[fill=white, rounded corners] (22.5,6.6) rectangle (25.5,8);
		\node[font=\scriptsize] at (24,7.3) {$\Psi$};
		\draw[->] (22.5,7.3) -- (21.5,7.3);
		\draw[fill=white] (21,7.3) circle (0.5);      
		\draw[fill=black] (22.222,4.5) circle (0.5);
		\draw[->] (23,6.6) -- (22.396,4.969);
		\draw[fill=black] (24,4.5) circle (0.5);
		\draw[->] (24,6.6) -- (24,5);
		\draw[fill=black] (25.778,4.5) circle (0.5);                
		\draw[->] (25,6.6) -- (25.64,4.969);
		\draw[fill=black] (29.333,9.5) circle (0.5);
		\draw[->] (25,11.6) -- (28.869,9.683);   
		\draw[->] (29.333,9) -- (29.333,8);
		\draw[fill=white, rounded corners] (27.833,6.6) rectangle (30.833,8);
		\node[font=\scriptsize] at (29.333,7.3) {$\Psi$};
		\draw[->] (27.833,7.3) -- (26.833,7.3);
		\draw[fill=white] (26.333,7.3) circle (0.5);
		\draw[fill=black] (27.555,4.5) circle (0.5);
		\draw[->] (28.333,6.6) -- (27.729,4.969);
		\draw[fill=black] (29.333,4.5) circle (0.5);
		\draw[->] (29.333,6.6) -- (29.333,5);
		\draw[fill=black] (31.111,4.5) circle (0.5);
		\draw[->] (30.333,6.6) -- (30.937,4.969);
		\draw[fill=black] (40,14.5) circle (0.5); 
		\draw[->] (25,16.6) -- (39.505,14.569);
		\draw[->] (40,14) -- (40,13);
		\draw[fill=white, rounded corners] (38.5,11.6) rectangle (41.5,13);
		\node[font=\scriptsize] at (40,12.3) {$\Psi$};
		\draw[->] (38.5,12.3) -- (37.5,12.3);
		\draw[fill=white] (37,12.3) circle (0.5);
		\draw[fill=black] (34.667,9.5) circle (0.5);
		\draw[->] (39,11.6) -- (35.132,9.683);
		\draw[->] (34.667,9) -- (34.667,8);
		\draw[fill=white, rounded corners] (33.167,6.6) rectangle (36.167,8);
		\node[font=\scriptsize] at (34.667,7.3) {$\Psi$};
		\draw[->] (33.167,7.3) -- (32.167,7.3);
		\draw[fill=white] (31.667,7.3) circle (0.5);        
		\draw[fill=black] (32.889,4.5) circle (0.5);
		\draw[->] (33.667,6.6) -- (33.063,4.969);
		\draw[fill=black] (34.667,4.5) circle (0.5);
		\draw[->] (34.667,6.6) -- (34.667,5);
		\draw[fill=black] (36.444,4.5) circle (0.5);
		\draw[->] (35.667,6.6) -- (36.27,4.969);
		\draw[fill=black] (40,9.5) circle (0.5);
		\draw[->] (40,11.6) -- (40,10);
		\draw[->] (40,9) -- (40,8);
		\draw[fill=white, rounded corners] (38.5,6.6) rectangle (41.5,8);
		\node[font=\scriptsize] at (40,7.3) {$\Psi$};
		\draw[->] (38.5,7.3) -- (37.5,7.3);
		\draw[fill=white] (37,7.3) circle (0.5);        
		\draw[fill=black] (38.222,4.5) circle (0.5);
		\draw[->] (39,6.6) -- (38.396,4.969);
		\draw[fill=black] (40,4.5) circle (0.5);
		\draw[->] (40,6.6) -- (40,5);
		\draw[fill=black] (41.778,4.5) circle (0.5);
		\draw[->] (41,6.6) -- (41.604,4.969);        
		\draw[fill=black] (45.333,9.5) circle (0.5);   
		\draw[->] (41,11.6) -- (44.833,9.683);
		\draw[->] (45.333,9) -- (45.333,8);
		\draw[fill=white, rounded corners] (43.833,6.6) rectangle (46.833,8);
		\node[font=\scriptsize] at (45.333,7.3) {$\Psi$};
		\draw[->] (43.833,7.3) -- (42.833,7.3);
		\draw[fill=white] (42.333,7.3) circle (0.5);        
		\draw[fill=black] (43.555,4.5) circle (0.5);
		\draw[->] (44.333,6.6) -- (43.729,4.969);
		\draw[fill=black] (45.333,4.5) circle (0.5);
		\draw[->] (45.333,6.6) -- (45.333,5);
		\draw[fill=black] (47.111,4.5) circle (0.5);
		\draw[->] (46.333,6.6) -- (46.937,4.969);
		
		\draw[fill=black] (24,2.9) circle (0.1);		
		\draw[fill=black] (24,2.5) circle (0.1);		
		\draw[fill=black] (24,2.1) circle (0.1);				
		\end{tikzpicture}
	\end{center}
	\caption{Scattering tree produced by successive application of $\Psi$ on the input signal $x$. Lowpass signals are depicted as white nodes. Once the tree is computed, only the white nodes are kept as the representation of the input signal.}
	\label{fig:WST_tree}
\end{figure}
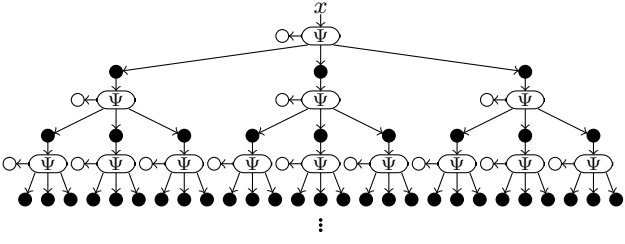
Simply put, a Scattering transform \cite{Mallat2012} is a convolutional neural net (CNN) with fixed weights and without channel recombination, where the absolute value operator is consistently used as the activation function. Consider an operator
$\Psi^J:L^2\to {L^2}^{J+1}$
that creates a subband decomposition of the input, \emph{and} applies the absolute value operator to each non-lowpass output. The first element of the output tuple, written as $\Psi^J[x]_{0}$ in the following, is a lowpass representation of the input. The other signals $\Psi^J[x]_{1},\dots,\Psi^J[x]_{J}$ contained in the tuple can be subjected to the operator again, yielding another $J$ subband decompositions with the absolute value applied to the bandpass signals. This procedure can be repeated several times, yielding a tree structure like in \Figref{fig:WST_tree}, where the black circles denote the absolute values of bandpass signals and the white circle denote lowpass signals. 
The $M$-depth Scattering transform of a signal $x$ is the collection of (only) the lowpass signals created by constructing a tree with $M$ layers. These signals are referred to \emph{subbands} in the following. Each Scattering subband of $x$ has the form
\begin{equation}
\mathcal{S}_{j_1,\dots, j_m}[x]=\Psi^{J_{m+1}}[\Psi^{J_m}[\cdots[\Psi^{J_1}[x]_{j_1}]\cdots]_{j_{m}}]_{0},
\end{equation}
with $m\in\{1,\dots,M-1\}$ and $j_m\in\{1, \dots, J_m\}$. The Scattering transform of $x$ results in a tuple containing all $N_\mathrm{Bands}=1+\sum_{m=1}^{M-1}\prod_{i=1}^{m}J_i$
subbands. For the subband at the top layer ($m=0$), let us fix the notation $\mathcal{S}[x]=\Psi^{J_1}[x]_{0}$.
As an optional step, the Scattering subbands can be normalized \cite{Sagel2015, Anden2014} in order to decorrelate them. Let us define the \emph{normalized Scattering transform} as a collection of subbands $\bar {\mathcal S}_{j_1,\dots, j_m}$ with
\begin{equation}
\bar{\mathcal S}[x]={\mathcal S}[x],\qquad
\bar{\mathcal S}_{j_1}[x]={{\mathcal S_{j_1}}[x]}/{\operatorname{avg}(x)},
\end{equation}
where $\operatorname{avg}$ denotes the signal average, and  
\begin{equation}
\bar{\mathcal S}_{j_1,\dots,j_m}[x]={\mathcal S_{j_1,\dots,j_m}[x]}/{\mathcal S_{j_1,\dots,j_{m-1}}[x]}
\end{equation}
for $m>1$.

In \cite{Sagel2015}, a texture retrieval method based on Scattering subband distributions has been proposed, in which texture images are described by the Weibull distribution parameters computed from the coefficients of each subband via maximum likelihood. We can thus conclude that the subband-wise distributions capture distinctive features from image textures.
The framework presented in the following is based on the same feature extraction mechanism, but the distributions are modeled as histograms.
Given a dynamic texture sequence $\vx_1,\dots\vx_N$ of vectorized video frames, let us assume that we have applied the Scattering transform to each one of the frames and computed a histogram of $N_\mathrm{bins}$ from each subband. To describe a video, we are thus given a matrix
\begin{equation}
\mH=\begin{bmatrix}
\vh_1 & \cdots & \vh_N
\end{bmatrix}\in \R^{N_{\mathrm{Bands}} N_\mathrm{bins}\times N},
\end{equation}
with Scattering histogram vectors as its columns, where $N_{\mathrm{Bands}}$ is the number of Scattering subbands.

\section{Representation via Kernel Subspaces}
\label{sec:ncd}
Assuming that the  Scattering histogram vectors are an appropriate way to represent the individual frames, the entire set of the columns in $\mH$ should capture the essential information of the whole dynamic texture, if we neglect the dynamics and the order of the frames. What we aim for is thus a compact way to represent this set. However, the classical approach of performing a principal component analysis (PCA) and representing $\mH$ by the principal subspace would likely fail, because low-dimensional linear vector spaces are not a fitting model for sets of histograms, which is why histogram data is often mapped to a feature space using the kernel trick, prior to further processing \cite{Chaudhry2009}.
Following the insights from \cite{Sagel2015}, we employ the \emph{Bhattacharyya} kernel which, for two probability distributions $p_1, p_2$ over the sample space $\Omega$, is defined as
\begin{equation}
\kappa_{\mathrm{Bht}}(p_1, p_2)=\int_{\Omega}\sqrt{p_1(\vomega)p_2(\vomega)}\operatorname{d}\vomega.
\end{equation}
Let us make the (simplifying) assumption that the coefficient distributions of different Scattering subbands are independent. Then we can evaluate the Bhattacharyya kernel on a pair of Scattering histogram vectors $\veta_1, \veta_2$ by computing
\begin{equation}
\kappa({\veta_1}, {\veta_2})=\prod_{i=0}^{N_{\mathrm{Bands}}-1}\sum_{j=1}^{N_\mathrm{bins}}\sqrt{{\veta_1}_{(i N_\mathrm{bins}+j)} {\veta_2}_{(i N_\mathrm{bins}+j)}}.
\end{equation}
For the sake of readability, we fix the notation
\begin{equation}
\kappa:\R^{N_{\mathrm{Bands}}N_\mathrm{Bins}\times N_1}\times \R^{N_{\mathrm{Bands}}N_\mathrm{Bins}\times N_2}\to\R^{N_1\times N_2}
\end{equation}
with $\kappa(\mH_1, \mH_2)_{(i,j)}=\kappa({\mH_1}_{(:,i)}, {\mH_2}_{(:,j)})$, for the matrix-wise application of $\kappa$.

Using $\kappa$ and $\mH\in\R^{N_\mathrm{Bands}N_\mathrm{Bins}\times N}$, we could now proceed to apply Kernel PCA to compute a low-dimensional subspace of a feature space corresponding to $\kappa$. Kernel PCA is typically carried out by performing truncated Eigenvalue or Singular Value Decomposition (SVD) of the \emph{Gram matrix} $\kappa(\mH, \mH)$ and returns a coefficient matrix $\mC\in\R^{N\times n}$ that, together with $\kappa$ and $\mH$, parameterizes an orthogonal basis spanning the $n$-dimensional principal subspace of the feature space representation of $\mH$ \cite{Schoelkopf1998}. The orthogonality of a basis described by a parameter pair $\mC,\mH$ can be easily verified using the condition
\begin{equation}
\mC^\top\kappa(\mH,\mH)\mC=\mI_n.
\label{eq:orthogonality}
\end{equation}
However, storing the whole matrix $\mH$ for each video sequence  is rarely sustainable.
So in addition to performing Kernel PCA, we include a \emph{Nystr{\"o}m interpolation} \cite{Drineas2005} step to approximate the computed subspace by a small subset of $\tilde{N}$ columns of $\mH$, for which the inequality $n\leq\tilde N < N$ holds. To do so, we collect $\tilde N$ representative columns from $\mH$ in a matrix $\tilde \mH \in \R^{N_\mathrm{Bands}N_\mathrm{Bins}\times \tilde N}$ and choose the coefficient matrix $\tilde\mC$, such that an appropriate metric between $\mC, \mH$ and $\tilde\mC, \tilde\mH$, is minimized. We choose the metric
\begin{equation}
\begin{split}
\frac{1}{2}\left(\tr(\mC_1^\top\kappa(\mH_1,\mH_1)\mC_1)+\tr(\mC_2^\top\kappa(\mH_2,\mH_2)\mC_2)\right)\\-\tr(\mC_1^\top\kappa(\mH_1,\mH_2)\mC_2)
\end{split}
\end{equation}
which measures the squared error between the basis vectors in the feature space. Since we can assume that both involved bases are orthogonal, i.e. \eqref{eq:orthogonality} holds for $\mC_1. \mH_1$ and $\mC_2. \mH_2$, we can write it as
\begin{equation}
\begin{split}
d_\mathrm{se}(\mC_1, \mH_1;\mC_2, \mH_2)^2=n-\tr(\mC_1^\top\kappa(\mH_1,\mH_2)\mC_2).
\end{split}
\label{eq:dse}
\end{equation}
We thus define
\begin{equation}
\tilde \mC = \argmin_{{
\mC'\mathrm{\ s.t.\ }\mC'^\top\kappa(\tilde \mH, \tilde \mH)\mC' = \mI_n}}d_\mathrm{se}((\mC, \mH),(\mC', \tilde \mH))^2.
\label{eq:trace_product}
\end{equation}
Due to \eqref{eq:orthogonality}, the solution of \eqref{eq:trace_product} must have the form 
$\tilde \mC = \tilde \mU\tilde\mLambda^{-1/2}\mW$,
where  $\tilde \mU, \tilde\mLambda$ denote the SVD factors of $\kappa(\tilde \mH, \tilde \mH)$ and $\mW\in\R^{\tilde N\times n}$ is a matrix with orthogonal columns, i.e. $\mW^\top\mW=\mI_n$. \eqref{eq:trace_product} thus boils down to
\begin{equation}
\hat \mW = \argmax_{{\mW\ \mathrm{s.t.}\ \mW^\top\mW=\mI_n}}\tr(\mC^\top\kappa( \mH, \tilde \mH)\tilde \mU\tilde\mLambda^{-1/2}\mW),
\label{eq:trace_prodcut2}
\end{equation}
which can be solved using the SVD.
\begin{algorithm}
	\KwIn{Video Sequence $\mX$, Histogram size $N_\mathrm{Bins}$, Sampling size $\tilde N$, Traget dimension $n$}
	\caption{\label{alg:scathist} Kernel Subspace Computation}
	$\mS \leftarrow \operatorname {ST}(\mX)$\tcp*{Scattering transform}
	$\mH \leftarrow \mathrm{hist}(\mS)$ \tcp*{Subband histograms}
	$\tilde{\mH}\leftarrow \begin{bmatrix}
	\vh_{i_1} & \cdots & \vh_{i_{\tilde N}}
	\end{bmatrix}$ \tcp*{Subsampling}
	$\mU, \mLambda, \mU\leftarrow \operatorname{SVD}(\kappa(\mH, \mH))$;\\
	$\mC \leftarrow \mU_{(:,1:n)}\mLambda_{(1:n,1:n)}^{-1/2}$\tcp*{Kernel PCA}
	$\tilde{\mU}, \tilde{\mLambda}, \_\leftarrow
	\operatorname{SVD}(\kappa(\tilde{\mH}, \tilde\mH))$;\\
	$\mU', \_, \mV'\leftarrow \operatorname{SVD}(\mC^\top\kappa(\mH, \tilde \mH)\tilde{\mU}\tilde{\mLambda}^{-1/2})$;\\
	$\tilde\mC \leftarrow \tilde{\mU}\tilde{\mLambda}^{-1/2}\mV'_{(:, 1:n)}\mU'^\top$\tcp*{\eqref{eq:trace_prodcut2}}
	\KwOut{Kernel Subspace parameters $\tilde \mC, \tilde{\mH}$}	
\end{algorithm}
Computing the subspace parameters of a dynamic texture is described in \Algref{alg:scathist}.

\section{Nuclear Distance on Kernel Subspaces}
Note that the parameters computed by \Algref{alg:scathist} are not unique. Consider a parameter pair $\mC, \mH$ describing an orthogonal basis in the feature space. Then, all pairs in the set
\begin{equation}
\{(\mC\mQ, \mH)\ |\ \mQ\in\operatorname{O}(n)\}
\label{eq:equiclass}
\end{equation}
describe \emph{different} orthogonal bases of the \emph{same} space. A semantically meaningful distance measure on kernel subspaces should be invariant to orthogonal transformations of $\mC$. This is not the case for \eqref{eq:dse}: Given a pair $\mC,\mH$, we observe
\begin{equation}
d_\mathrm{se}^2(\mC, \mH; -\mC, \mH)=2n>0=d^2_\mathrm{se}(\mC, \mH; \mC, \mH),
\end{equation}
even though both $\mC, \mH$ and $-\mC, \mH$ pairs describe the same subspace. We overcome this ambiguity by choosing $\mQ$ always such that $d^2_\mathrm{se}$ is minimized. We denote the result 
\begin{equation}
\min_{\mQ\in\operatorname{O}(n)} d^2_\mathrm{se}(\mC_1, \mH_1; \mC_2\mQ,\mH_2)
\end{equation}
the kernelized \emph{Nuclear distance}, as it is obtained by replacing the trace in \eqref{eq:dse} by the nuclear norm. On two matrix pairs $\mC_1, \mH_1;$ $\mC_2, \mH_2$, the Nuclear distance is computed as
\begin{equation}
\begin{split}
\dncl^2(\mC_1, \mH_1; \mC_2,\mH_2)=n-\|\mC_1^\top\kappa(\mH_1,\mH_2)\mC_2\|_*.
\end{split}
\label{eq:nuclear}
\end{equation}
\eqref{eq:nuclear} is actually a special case of the kernelized \emph{Alignment distance} \cite{Afsari2012,Sagel2017}. As such, it inherits its metric property on the the equivalence classes in \eqref{eq:equiclass}. Specifically, $d_\mathrm{ncl}$, i.e. the square root of \eqref{eq:nuclear} is symmetric, positive definite and fulfills the triangle inequality \cite{Sagel2017}.
\section{Fr{\'e}chet Means as Abstract Averages}
\label{sec:frechet}
Certain scenarios, such as clustering or \emph{nearest class center} (NCC) classification require the notion of an average for finite sets of descriptors.  Sadly, the set of kernel subspaces with dimension $n$ is a non-trivial manifold, which means we can not compute a mean using the arithmetic average. Thanks to \eqref{eq:nuclear}, we are operating on a metric space, which means that we can define averages using the \emph{Fr{\'e}chet mean}.
Consider a set $\{(\mC_i, \mH_i)\}_{i\in\{1,\dots,K\}}$. Its Fr{\'e}chet mean  is given by
\begin{equation}
\bar{ \mC}, \bar{ \mH}=\argmin_{\substack{\mC, \mH\mathrm{\ s.t.}\\ \mC^\top\kappa(\mH, \mH)\mC = \mI_n}} \sum_{i=1}^{K}\dncl^2(\mC, \mH;\mC_i, \mH_i).
\label{eq:frechetloss}
\end{equation}
To solve the optimization problem, we follow the approach in \cite{Sagel2017}. We approximate $\bar{ \mH}$ by computing $\bar N$ cluster centers using the  $k$-means algorithms on the columns of $\mH_1,\dots, \mH_K$. Next,  $\bar \mC$ is approached using an alternating scheme with two steps for each iteration.
The first step consists of finding a set of orthogonal matrices $\mQ_1,\dots,\mQ_K$, such that
\begin{equation}
\dncl^2(\bar\mC, \bar\mH;\mC_i, \mH_i)=d_\mathrm{se}^2(\bar\mC, \bar\mH;\mC_i\mQ_i, \mH_i),
\label{eq:qi_update}
\end{equation}
is fulfilled, for every $i\in\{1,\dots,K\}$. This can be easily achieved using a projection onto the orthogonal group. The second step is approximating $\bar C$ by computing
\begin{equation}
\argmin_{\mC\ \mathrm{\ s.t.\ }\mC^\top\kappa(\bar\mH, \bar\mH)\mC = \mI_n} \sum_{i=1}^{K}d_\mathrm{se}^2(\mC, \bar\mH;\mC_i\mQ_i, \mH_i).
\label{eq:se_update}
\end{equation}
These two steps are repeated until convergence of the loss function in \eqref{eq:frechetloss}. Due to space constraints, we omit providing pseudo-code for the final algorithm and  refer the reader to \cite{Sagel2017} for in-detail description of a similar algorithm.

\section{Experiments}
The DynTex database \cite{Peteri2010} is a collection of high-resolution  RGB  texture  videos.  Three  splits have been compiled for classification benchmarking.

\emph{DynTex  Alpha} is  composed  of  60 videos divided into the 3 classes Sea (20 videos), Grass (20), and Trees (20)
 
\emph{DynTex  Beta} is  composed  of  162  videos divided into the 10 classes Sea (20), Vegetation (20), Trees (20), Flags (20), Calm Water (20), Fountains (20), Smoke (16), Escalator (7), Traffic (9) and Rotation (10).

\emph{DynTex  Gamma} is  composed  of  264  videos divided into the 10 classes Flowers (29), Sea (38), Naked  trees (25), Foliage (35), Escalator (7), Calm water (30), Flags (31), Grass (23), Traffic (9) and Fountains (37). Some works \cite{Dubois2015} use a different version of this split containing 275 videos.

We perform $1$-NN (Nearest Neighbor) and NCC classification experiments on the three splits. To this end, \emph{Kernelized Scattering Histogram Spaces} (KSHS) are computed from each video using \Algref{alg:scathist} with  $n=5, \tilde N =15$. The Scattering transform is computed using Kymatio \cite{Andreux2020} on CUDA with the arguments \texttt{L=4, J=4}. Histograms with $N_\mathrm{Bins}=20$ are calculated, from both regular and normalized (KNSHS) subbands. For NCC classification, a Fr{\'e}chet mean is computed from each class using the procedure described in \Secref{sec:frechet} with $\bar N=15$ and the leave-one-out protocol proposed in \cite{Dubois2015}. Parameters were chosen using grid search. Code for reproducing the experimental results is available online\footnote{https://github.com/alexandersagel/kshs}.
\begin{table}
	\scriptsize\begin{center}
		\hspace*{-5pt}
		\begin{tabular}{c|cc|cc|cc}
			& \multicolumn{2}{c|}{Alpha}&\multicolumn{2}{c|}{Beta}&\multicolumn{2}{c}{Gamma} \\ 
			& $1$-NN&NCC&$1$-NN&NCC&$1$-NN&NCC \\ \hline\hline
			LBP-TOP &96.7\ \% &-& 85.8\ \% &-& 84.9\ \%&-\\ \hline
			PCANet-TOP&96.7 \%& -&90.7 \%& -&89.4 \%& - \\ \hline
			DFS &-&  83.6 \%&-& 65.2 \%& - &60.8  \%\\ \hline
			2D+T C.&- &85.0 \%  &-&67.0 \% &-&-* \\ \hline
			OTDL&-&86.6 \%&-&69.0 \%& -&64.2 \% \\ \hline
			CLSP-TOP&95.0 \%&-&92.0 \%&-& \textbf{91.3} \%& - \\ \hline
			STRF N-jet  &\textbf{100.0 \%}  &-&93.8\%  &-& 91.2 \%&- \\ \hline
			B3DF& 96.7 \%& 90.0 \%&90.1 \%& 74.1 \%&-*&-* \\ \hline
			SOE-NET &98.3 \% &\textbf{96.7 \%}&\textbf{96.9 \%}&86.4 \%&-* &-* \\ \hline
			SoB+Align &98.3 \%& {88.3 \%} &90.1 \%& 75.3 \%&79.9 \%& {67.1 \%} \\ \hline\hline
			KSHS+Ncl.  &98.3 \%& \textbf{96.7 \%} &88.9 \%& 88.3 \%&88.6 \%& {86.7 \%} \\ \hline
			KNSHS+Ncl. &98.3 \%& \textbf{96.7 \%} &93.2 \%& \textbf{90.1 \%}&\textbf{91.3} \%& \textbf{89.8 \%}
		\end{tabular}
	\end{center}
	\vspace{-6pt}\tiny{*Reported results refer to 275-video version of DynTex Gamma.}	
	\caption{Classification rate on DynTex subsets}
	\label{tab:dyntex_st}
\end{table}

\Tabref{tab:dyntex_st} shows the 1-NN and NCC classification results in comparison with 
LBP-TOP \cite{Zhao2007,Qi2016}, PCANet-TOP \cite{Arashloo2017},
Dynamic  Fractal  Spectrum (DFS) \cite{Xu2011, Quan2015}, Spatiotemporal Curvelet Transform (2D+T Curvelet) \cite{Dubois2015}, Orthogonal Tensor Dictionary Learning (OTDL) \cite{Quan2015}
Completed Local Structure Patterns in Three Orthogonal Planes (CLSP-TOP)  \cite{Nguyen2017}, Spatio-temporal Receptive Fields (STRF N-jet) \cite{Jansson2018}, Binarized 3D  features (B3DF) \cite{Zhao2019} Spatiotemporal Oriented Energy Network (SOE-NET) \cite{Hadji2017}, and Systems of Bags with the Alignment Distance (SoB+Align) \cite{Sagel2017}. Results that have not been reported or are not applicable are indicated by '-'.

In line with the results in \cite{Sagel2015}, normalizing the Scattering coefficients improves the recognition performance: for no experimental setting, KNSHS performs worse than KSHS. One explanation is that normalized coefficients are closer to fulfilling the independence assumption made in \Secref{sec:ncd}.
Overall, KNSHS in combination with the Nuclear distance competes well with current approaches. For the $1$-NN classification, it yields the same success rate as CLSP-TOP on the Gamma split, and is outperformed by STRF N-jet and SOE-NET on the Alpha and Beta split of DynTex.
More remarkable are the results for NCC classification. Not only does our method yield higher success rates than many of the state-of-the-art approaches in literature, but it does so with a considerably small performance gap with regards to $1$-NN classification. This could be due to using Fr\'echet means as class centers: Because of the triangle inequality, the distance from a test point to the Fr\'echet mean of a set is always a good approximation of the distance from the test point to any point in said set \cite{Sagel2017}. Hence, NCC should yield similar results as $1$-NN.
\section{Conclusion}
In this work, we have proposed a Scattering-based feature extraction method for dynamic textures using Kernel PCA, and described a {distance} that accounts for non-uniqueness of the extracted features. Additionally, we have briefly outlined a procedure to compute abstract averages from finite sets of such features.
We have evaluated the proposed method on $1$-NN and NCC classification and have observed state-of-the-art results for the latter scenario.
The capability of computing expressive features and measure the distance between pairs thereof, in addition to computing abstract averages, are also useful in related recognition tasks, such as retrieval, clustering or even segmentation of video data.
\pagebreak
\bibliographystyle{IEEEbib}
\bibliography{refs}

\end{document}

%% file: math_commands.tex
\usepackage{amsmath,amsfonts,bm,amsthm}

\def\Figref#1{Figure~\ref{#1}}

\def\Secref#1{Section~\ref{#1}}

\def\eqref#1{Eq.~(\ref{#1})}

\def\Algref#1{Algorithm~\ref{#1}}

\def\Tabref#1{Table~\ref{#1}}

\def\1{\bm{1}}

\def\veta{{\bm{\eta}}}
\def\vomega{{\bm{\omega}}}

\def\vh{{\bm{h}}}

\def\vx{{\bm{x}}}

\def\mC{{\bm{C}}}

\def\mH{{\bm{H}}}
\def\mI{{\bm{I}}}

\def\mQ{{\bm{Q}}}

\def\mS{{\bm{S}}}

\def\mU{{\bm{U}}}
\def\mV{{\bm{V}}}
\def\mW{{\bm{W}}}
\def\mX{{\bm{X}}}

\def\mLambda{{\bm{\Lambda}}}

\DeclareMathAlphabet{\mathsfit}{\encodingdefault}{\sfdefault}{m}{sl}
\SetMathAlphabet{\mathsfit}{bold}{\encodingdefault}{\sfdefault}{bx}{n}

\newcommand{\tr}{\operatorname{tr}}

\newcommand{\R}{\mathbb{R}}

\DeclareMathOperator*{\argmax}{arg\,max}
\DeclareMathOperator*{\argmin}{arg\,min}